\documentclass{isprs} 
\usepackage{subfigure}
\usepackage{setspace}
\usepackage{geometry} 
\usepackage{epstopdf}
\usepackage[labelsep=period]{caption}  
\usepackage[british]{babel} 
\usepackage[hang]{footmisc}
\usepackage{amsmath}
\usepackage{amssymb}
\usepackage{xcolor}

\newcommand{\topfivewidth}{0.26\textwidth}
\newcommand{\kittifivewidth}{0.70\textwidth}
\newcommand{\abwidth}{0.3\linewidth}

\usepackage[authoryear]{natbib}

\usepackage{balance}

\usepackage{xcolor}
\usepackage[normalem]{ulem}

\begin{document}

\title{Evaluation of Visual Place Recognition Methods for
Image Pair Retrieval in\\ 3D Vision and Robotics}
\date{}


\author{
Dennis Haitz\textsuperscript{1}\thanks{Corresponding author}, 
Athradi Shritish Shetty\textsuperscript{1}, 
Michael Weinmann\textsuperscript{2}, 
Markus Ulrich\textsuperscript{1}
}

\address{
\textsuperscript{1}Karlsruhe Institute of Technology, Institute of Photogrammetry and Remote Sensing, Karlsruhe, Germany - \\(dennis.haitz, markus.ulrich)@kit.edu, athradi.shetty@student.kit.edu \\
\textsuperscript{2} Delft University of Technology, Department of Intelligent Systems, The Netherlands  - \\
    m.weinmann@tudelft.nl 
}



\abstract{
Visual Place Recognition (VPR) is a core component in computer vision, typically formulated as an image retrieval task for localization, mapping, and navigation. In this work, we instead study VPR as an \textit{image pair retrieval} front-end for registration pipelines, where the goal is to find top-matching image pairs between two disjoint image sets for downstream tasks such as scene registration, SLAM, and Structure-from-Motion. We comparatively evaluate state-of-the-art VPR families - NetVLAD-style baselines, classification-based global descriptors (CosPlace, EigenPlaces), feature-mixing (MixVPR), and foundation-model-driven methods (AnyLoc, SALAD, MegaLoc) - on three challenging datasets: object-centric outdoor scenes (Tanks and Temples), indoor RGB-D scans (ScanNet-GS), and autonomous-driving sequences (KITTI). We show that modern global descriptor approaches are increasingly suitable as off-the-shelf image pair retrieval modules in challenging scenarios including perceptual aliasing and incomplete sequences, while exhibiting clear, domain-dependent strengths and weaknesses that are critical when choosing VPR components for robust mapping and registration. 
}


\maketitle


\section{Introduction}\label{sec:introduction}
\sloppy
Visual place recognition (VPR) is a computer vision task that seeks to retrieve similar images of the same scene based on a query image. 
While image retrieval is a task that solely relies on information derived from the image itself, VPR methods often further utilize the acquisition position, i.e.\ images are assumed to be georeferenced or geotagged. 
From a set of search images, the top-$k$ most similar images w.r.t.\ the query image are to be retrieved, where the $k$ search images ideally show the same scene content as the query image. 
Challenges in VPR are different viewpoints and viewing angles, illumination and weather changes, or even object manipulations or displacements, e.g. through moving cars at different acquisition times. Additionally, perceptual aliasing plays a role, including recurring texture or patterns through the scene or sequence.
Typical use-cases of VPR are large-scale outdoor scenes, e.g. in urban and rural environments.\\
While the performance of early CNN-based approaches is impacted by their limited receptive fields, the introduction of Vision transformers (ViT) with their attention mechanisms allows to better capture long-range dependencies.
A wide range of the state-of-the-art (SOTA) methods are now built on top of DINOv2~\citep{oquab2023dinov2} as a ViT-based visual feature encoder.\\
In this contribution, we put our attention on investigating the suitability of different conceptual approaches for finding a fast and reliable association between two sets of RGB images with overlapping scene content, without taking further availability of geometry data into account to avoid further computational overhead.
This is of great relevance for the registration of RGB-D images or 3D Gaussian Splatting (3DGS) models \citep{Kerbl_et_al_2023}  with overlapping scene content, where the image sets or the 3DGS models may not initially be located in the same coordinate system and a georeferencing, e.g. a coarse image acquisition position with heading direction assumed through geographic or UTM coordinates, in the same coordinate system is not available.
In order to tackle this task, geometric approaches can be applied, where depth maps can be projected into 3D space based on known camera parameters to get a point cloud.
This allows applying known point cloud registration methods to receive an initial coarse registration, in the form of an estimated transformation matrix.
While methods based on iterative-closest-point (ICP) \citep{Besl1992_ICP} are often prone to local minima, they cannot establish the relative pose if point clouds do not overlap without further computation efforts.\\
\setlength{\fboxsep}{0pt}
\setlength{\fboxrule}{1.5pt}
\begin{figure}[ht]
  \centering
  \setlength{\tabcolsep}{1pt}
  \renewcommand{\arraystretch}{1.0}
  \begin{tabular}{cc}
    A & B \\[1pt]
    \includegraphics[width=\abwidth]{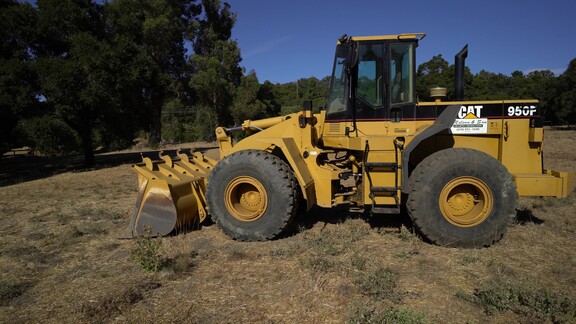} &
    \fcolorbox{red}{white}{\includegraphics[width=\abwidth]{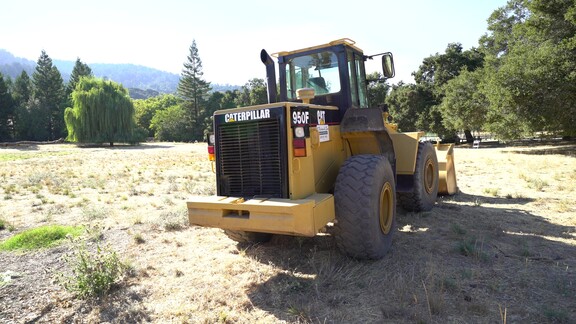}} \\
    \includegraphics[width=\abwidth]{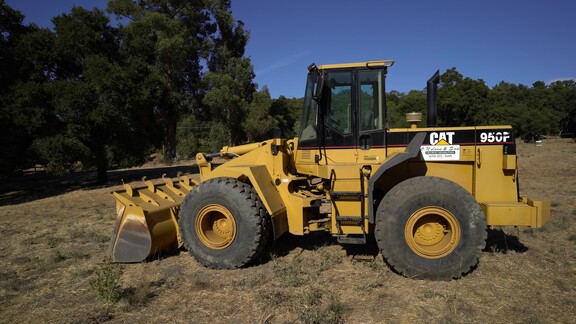} &
    \fcolorbox{red}{white}{\includegraphics[width=\abwidth]{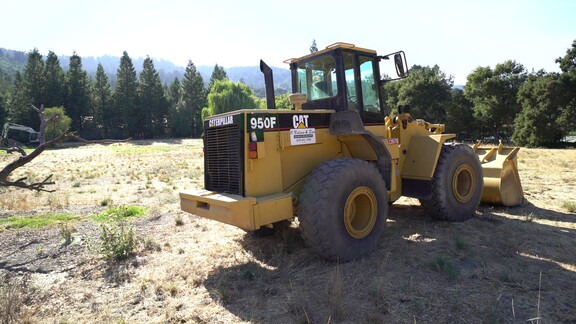}} \\
    \includegraphics[width=\abwidth]{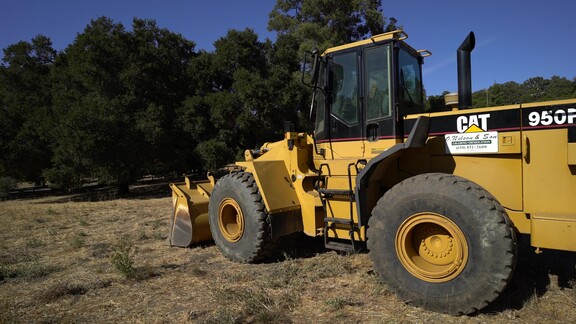} &
    \includegraphics[width=\abwidth]{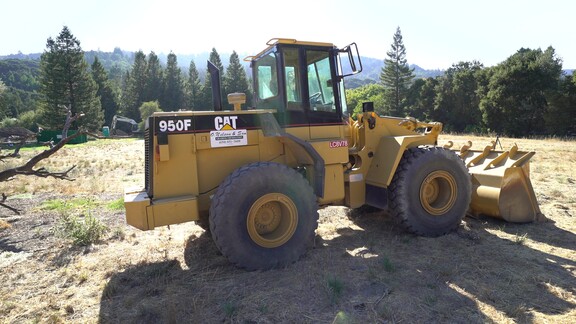} \\
    \fcolorbox{red}{white}{\includegraphics[width=\abwidth]{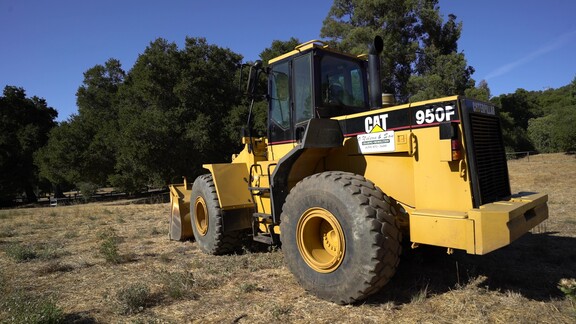}} &
    \includegraphics[width=\abwidth]{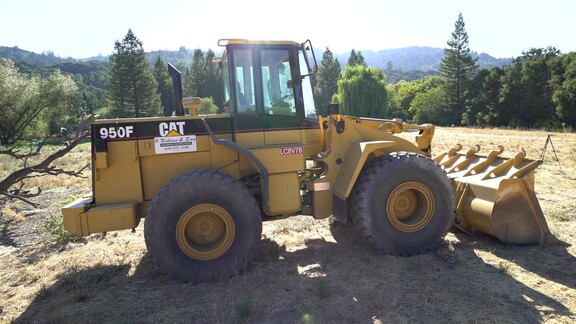} \\
    \fcolorbox{red}{white}{\includegraphics[width=\abwidth]{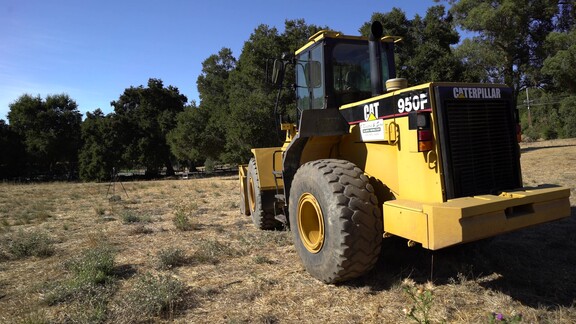}} &
    \includegraphics[width=\abwidth]{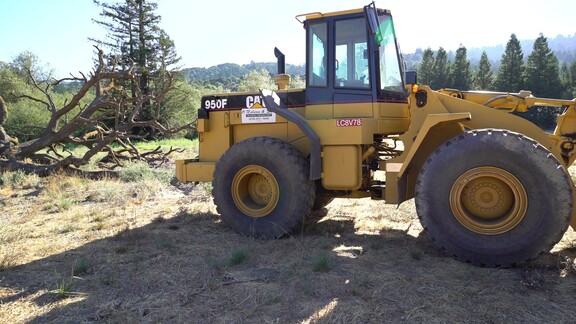} \\
    \fcolorbox{red}{white}{\includegraphics[width=\abwidth]{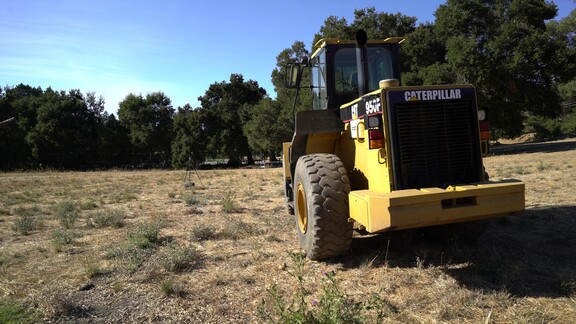}} &
    \includegraphics[width=\abwidth]{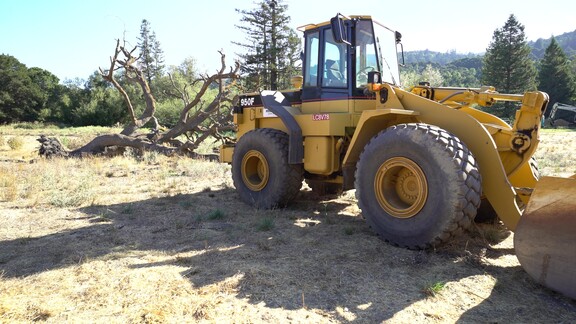} \\
  \end{tabular}
  \caption{Example images from sets A (left) and B (right), taken from the Caterpillar scene of the Tanks and Temples dataset~\citep{Knapitsch2017}. Images including overlapping scene area are framed in red, the objective in our contribution is to test the ability of VPR methods to retrieve these pairs. Challenging for VPR models in such scenes is the perceptual aliasing, e.g. indicated by the wheels, which are visible on both sides and even appear in similar relative positions.}
  \label{fig:tt-caterpillar-AB}
\end{figure}
%
Regarding runtime efficiency, deep-learning-based methods can process hundreds of images within few seconds with GPU support. For image sequences, the search space can further be reduced if keyframes are taken into account.
In the case of RGB-D registration, stereo-camera setups with known intrinsic and extrinsic camera parameters can be considered as potential application scenarios. 
Regarding 3DGS, cameras with intrinsics and extrinsics are already assumed to be known beforehand, in order to be able to train a 3DGS model. 
The RGB images for image-driven registration can then be rendered through given extrinsics from such models. 
Besides common indoor and outdoor datasets, we want to evaluate VPR methods on challenging scenes with ambiguous content, e.g. repeating textures or objects. 
Those scenes are often captured with object-centric trajectories. 
We prepare scenes of the Tanks-and-Temples~\citep{Knapitsch2017} dataset for this specific case. 
In addition, we also focus on indoor and outdoor setups, with ScanNet-GSReg~\citep{Chang:2024} and KITTI-odometry~\citep{Geiger2012CVPRkitti} as the base for our prepared datasets, respectively.
Our evaluation is therefore set apart from the usual VPR training and test case, which includes large-scale data. 
Only recently, MegaLoc~\citep{Berton_2025_MegaLoc} took ScanNet as an indoor dataset into account, which is the superset of our used ScanNet-GSReg dataset.

Our contributions are the following:
\begin{itemize}
    \item An evaluation pipeline with scale-independent matching criteria for photogrammetric reconstructions, targeted at typical 3D vision and robotics applications
    \item Metrics that elevate standard-practice in VPR towards image pair matching
    \item Providing prepared subsets of well-known datasets (KITTI, Tanks and Temples) for the application of image-driven scene registration.
\end{itemize}

\section{Related Work}\label{sec:related-work}

%
In the following, we provide a brief survey of the developments regarding VPR.
%
\paragraph{Hand-crafted feature approaches.}
Early developments from the pre-deep-learning era focused on the representation of scene characteristics shown in images in terms of hand-crafted features. 
Representations based on \emph{global features} such as GIST descriptors~\citep{Oliva:2001:GIST} and their respective variants~\citep{Murillo:2013} aggregate the gradient characteristics of an image in a single vector representation to capture the holistic scene structure.
In contrast, other representations leverage sets of distinctive \emph{local features}, such as SIFT, SURF, and ORB descriptors that are invariant to scale and rotation.
The conversion of distributions of local descriptors into compact global descriptors, can be achieved based on computing bag-of-words representations~\citep{Sivic:2003:VideoGoogle,Csurka04visualcategorization,Nister:2006:VocabularyTree} or vectors of locally aggregated descriptors (VLAD)~\citep{Jegou:2010,Arandjelovic:2013} that both rely on the initial generation of codebook descriptors (e.g., based on K-means or mean-shift clustering) and a subsequent assignment of the individual descriptors to their closest codebook entry (i.e., word) to get a histogram-like representation or the aggregation of the offset vectors of the descriptors assigned to a particular codebook entry, respectively. 
Furthermore, sequence-based approaches~\citep{Milford:2012:SeqSLAM,Bampis:2016} leverage temporal image matching for place recognition, i.e. the matching of sequences of images rather than individual frames, thereby improving robustness to appearance changes.
While computationally efficient, these methods often fail under severe illumination or seasonal variations as well as significant changes of view conditions.

\paragraph{Deep-feature approaches.}
With the advent of deep learning, representations via CNN features (e.g., obtained from networks trained on large datasets like ImageNet) were demonstrated to outperform handcrafted descriptors for landmark retrieval~\citep{Razavian:2014,Babenko:2014}.
Inspired by VLAD-based image representations based on hand-crafted local features, NetVLAD~\citep{ArandjelovicGTP16_NetVLAD} introduced a trainable VLAD layer to aggregate CNN features into a single robust image descriptor, whereas others introduced intermediate layer pooling~\citep{Chen:2017:AMOSNet+HybridNet} to exploit mid-level features to capture both semantic and structural cues.
Focusing on better pooling functions, GeM pooling (Generalized-Mean pooling) was introduced to unify average and max pooling and improve compact global descriptors~\citep{Radenovic:2019:GeMpooling}.
Whereas global descriptors are efficient but may miss fine geometric cues and local features are accurate but accompanied by higher computational demands, several works (e.g., ~\citep{Cao:2020:DELG,Yang:2021:DOLG}) focused on combining the advantages of both of these approaches within hybrid architectures that unify local and global representations.
%
%
%
%
%

\paragraph{Increased robustness to viewpoint variations and cross-domain conditions.}
To provide additional robustness for place recognition under varying viewpoints, recent developments particularly focused on multi-view training, patch-level descriptors, multi-scale fusion, and transformer backbones that leverage attention for region selection and robustness.
Examples include the training of lightweight CNNs with varied viewpoints~\citep{Khaliq:2020} and the training on multiple views of the same scene with a per-place clustering of images from diverse viewpoints into the same label to enforce the network to learn descriptors invariant to perspective changes~\citep{Berton_2023_EigenPlaces}.
%
Others leveraged local feature matching at retrieval time, e.g., in terms of performing dense correspondence matching after initial retrieval to verify matches across wide baselines~\citep{Berton:2021}.
Furthermore, PatchNetVLAD~\citep{hausler2021patchnetvlad} computes patch descriptors from NetVLAD residuals and fuses local and global descriptors at multiple scales to gain robustness against viewpoint changes.
Further work includes the use of context-flexible attention models to focus on stable landmarks for long-term place recognition~\citep{Chen:2018,Liu:2019} as well as the use of multi-level attention to highlight task-relevant regions and combine global and key-patch descriptors~\citep{Wang:2022:TransVPR}.
An alternative strategy to address viewpoint gaps consists in re-formulating the problem as an overlap prediction task, where spatial verification techniques are used to predict the overlap or relative pose between images to determine if they depict the same place~\citep{Wei:2025:OverlapPrediction}.\\
To address scalability issues of contrastive pairwise training and training based on triplet losses and hard negative mining for large training datasets, CosPlace~\citep{Berton_CVPR_2022_CosPlace} partitions the scene into many discrete "place classes" and trains a classifier to predict the place label from an image, leading to state-of-the-art results with significantly more compact descriptors and greatly reduced memory usage.
Building on that idea, MixVPR~\citep{Ali-bey_2023_WACV_MixVPR} used feature mixing as a form of data augmentation for VPR to further improve the generalization capabilities.\\
Another recent trend is the exploitation of foundation models for VPR, e.g. in terms of using a large pre-trained ViT (such as DINOv2) as a frozen feature extractor~\citep{keetha2023anyloc,Izquierdo_CVPR_2024_SALAD} or the fine-tuning of such foundation features for VPR~\citep{Izquierdo_CVPR_2024_SALAD}.
To improve the aggregation, SALAD~\citep{Izquierdo_CVPR_2024_SALAD} reformulates NetVLAD’s soft assignment as an optimal transport problem, introducing a Sinkhorn-based clustering (with a “dustbin” to drop non-informative features) to produce a more discriminative global descriptor.\\
Further improvements were achieved by cross-domain models that are applicable across diverse domains such as indoor and outdoor scenarios, day/night conditions, or seasonal changes.
Examples include the training of networks on large datasets of time-varying outdoor webcam and street-view data, leading to improved robustness to illumination and seasonal changes ~\citep{Chen:2017:AMOSNet+HybridNet,hausler2021patchnetvlad,Wang:2022:TransVPR}, as well as the use of unsupervised fine-tuning strategies and dataset-wise automatic supervision using 3D reconstructions or SfM to mine positives/negatives~\citep{Radenovic:2019:GeMpooling} that improved retrieval performance without manual annotation.
More recently, the training on a mix of datasets has also been followed with recent, matured models~\citep{Berton_2025_MegaLoc} that leverage previous insights on data augmentation, multi-domain training, and large-batch learning to handle different conditions with a single model, leading to robust performance and generalization for across autonomous driving scenes (KITTI), indoor scans (ScanNet), and object-centric datasets (e.g., Tanks and Temples).



\section{Methodology}\label{sec:methodology}
This section lays out the method for evaluating SOTA VPR methods as detectors for image pairs that show a common scene area. 
We first introduce common VPR methods, their evaluation procedure and lastly, present our evaluation setup for image pair retrieval using VPR.
Additionally, we reduce the problem of image pair retrieval to two image sets without loss of generality, as it can certainly be the case to find image pairs in multiple image sets, showing the same scene. 
Therefore, our setup can be considered as a subtask of a multi-dataset problem.


\subsection{Preliminaries for VPR}
Neural-network-based VPR methods are usually formulated as image retrieval problems, often in conjunction with runtime efficiency, included in the broad VPR literature and usually denoted as \textit{extraction time}~\citep{Berton_CVPR_2022_CosPlace, Berton_2023_EigenPlaces, Ali-bey_2023_WACV_MixVPR}.
In real-time scenarios like (online) SLAM, where VPR methods potentially can be used for loop-closure or relocalization, this is of high importance. 
Thus, it is necessary to utilize an image representation that yields high computational efficiency, e.g.\ realized through vectors.
%
Whereas the matching of vectors is a well-researched topic in the field of information retrieval, many VPR methods do not aim at fast vector matching but rather focus on obtaining robust representational vectors that preserve high matchability in challenging scenarios.
These scenarios include perceptual aliasing, large viewpoint distances, and large view-direction angles.
%
In the following, we discuss core aspects of VPR methods in the three categories of \textit{backbone and aggregation}, denoting general architecture, \textit{loss functions and training} and \textit{reranking}, which is usually the second part of two-stage approaches.

\paragraph{Backbone and Aggregation.}
Convolutional Neural Networks (CNNs) parameterized by their weights $\theta$ provide the possibility to extract local feature characteristics and derive a vector representation of an image \mbox{$f_{\theta}: I\mapsto v$}.
Early deep-learning-based methods like NetVLAD~\citep{ArandjelovicGTP16_NetVLAD} approached this by training a CNN to produce image representations $v$ that are directly targeted at VPR by minimizing the Euclidean distance between the representation of two images. 
The backbones of such methods were often pretrained ResNet models of different depth, usually depths of 18, 50, or 101. Beyond these options, VGG16~\citep{simonyan2015deepconvolutionalnetworkslargescaleVGG} can also be utilized in CosPlace~\citep{Berton_CVPR_2022_CosPlace} and EigenPlaces~\citep{Berton_2023_EigenPlaces} as the underlying architecture. 
Instead, more recent methods like SALAD~\citep{Izquierdo_CVPR_2024_SALAD}, MegaLoc~\citep{Berton_2025_MegaLoc} or AnyLoc~\citep{keetha2023anyloc} are built on ViTs.
Especially DINOv2~\citep{oquab2023dinov2}, which is pretrained on a large dataset of 142 million images and aims at providing meaningful domain-independent visual features, is often utilized as a feature extraction backbone.\\
Attached to the backbone is an aggregation strategy to obtain the global image representation or descriptor $v$ from local features, obtained from, for example, feature maps in CNNs or patch embeddings in ViTs. 
Aggregations consist of an often small amount of additional network layers, like fully-connected, convolution, or activation layers. 
Fully-connected and convolution layers also imply that a training is necessary if they are included in the aggregation. 
The NetVLAD layer~\cite{ArandjelovicGTP16_NetVLAD} has been widely adopted for aggregation. It learns the position of a pre-defined number of cluster centers at training time. 
The descriptor is then calculated from the residuals of the output of the backbone and the cluster centers, weighted by a soft-assignment to further refine the feature-assignment to a cluster-induced visual context.
Applying a Generalized Mean Pooling (GeM)~\citep{Radenovic:2019:GeMpooling} in addition to fully-connected and $L_2$-normalization layers is a further strategy for aggregation. 
GeM pools its input depending on a learnable parameter, which determines an interpolation between average and maximum pooling.

\paragraph{Loss functions.}
Contrastive and triplet losses are common for a lot of VPR approaches. The Euclidean distance of the descriptors of positive and negative samples are usually maximized through these losses. For triplet losses, an additional anchor is added as a third loss input. 
Descriptors as representations of images are defined as positives, negatives, or anchors either through labeling, based on georeferencing with distance and angular thresholds, or images of the same place at different times, inducing different visual cues.
On the other hand, methods like CosPlace~\citep{Berton_CVPR_2022_CosPlace} and EigenPlaces~\citep{Berton_2023_EigenPlaces} formulate their training procedure as a classification problem, utilizing a cross-entropy-based loss. Classes represent grid cells including places, where a place is represented through a point in a coordinate system. In case of CosPlace, this place stems from georeferencing and is further assigned to the grid cell, defined across all places at a certain resolution. To mitigate place ambiguities induced by similar places with vastly differing view directions per cell, each cell is further split into multiple classes, representing heading bins within a cell.

\paragraph{Reranking.}
Reranking is often added as a second stage within a VPR pipeline and has the objective of further refining the top results. PatchNetVLAD~\citep{hausler2021patchnetvlad} is a method that utilizes local cues from the descriptor building process to add a second stage. Because the feature vectors per position are downsampled and condensed representations of images, these vectors represent actual patches with additional information about their neighboring patches through convolution operations. This patch-based approach is also utilized within ViT pipelines, because a ViT creates patch embeddings, which consequently can be reused for reranking on a patch level~\citep{Zhang_2023_WACV_ETR}. However, methods like SALAD~\citep{Izquierdo_CVPR_2024_SALAD} or MixVPR~\citep{Ali-bey_2023_WACV_MixVPR} show and explicitly state that modern global-descriptor-only approaches outperform reranking, which often also add a significant computational overhead that is not feasible for real-time applications. The use of only global descriptors is further underlined by\cite{Shao_2023_ICCV}, who also argue that a lightweight refinement of global descriptors can be sufficient for superior results regarding local reranking.

\subsection{Preliminaries for VPR Evaluation}\label{sec:preliminaries-vpr-eval}
The common evaluation procedure in the VPR literature involves comparing a retrieved subset (retrieval set) of top-$k$ images from a search set, typically provided by an image database, to a query image. 
To classify an image in the retrieval set as positive w.r.t. the query image, different criteria emerged across different datasets and methods. 
Because images are often georeferenced in VPR scenarios, a true positive image within the retrieval set is defined by the metric viewpoint position distance from the query image to the retrieval image. 
This distance is set to 25 meters in evaluation scenarios across all methods for reasons of comparability, e.g. in~\citep{ArandjelovicGTP16_NetVLAD, Ali-bey_2023_WACV_MixVPR, Izquierdo_CVPR_2024_SALAD, Berton_2023_EigenPlaces}. 
With the introduction of the Mapillary dataset~\citep{Warburg2020_Mapillary_StreetLevel_CVPR}, an additional view direction difference of smaller than 40 degrees was taken into account for the ground truth definition of a match.
%
The Nordland dataset~\citep{sunderhauf2013we_Nordland} was acquired along a railroad from a train in different seasons, with the camera facing in driving direction. Though in the evaluation in MixVPR, the ground truth for this dataset is for the viewpoints of the images to lie within a distance of 25 meters,~\cite{Malone_2025_ICCV} set a number of frames two viewpoints may lie apart. Therefore, the sequential nature of such datasets can also be taken into account.\\
Retrieval performance is typically evaluated based on the 
\textit{recall@k} metric (R@k). This metric is calculated as the average of all retrievals (which is the number of all utilized query images), that contain at least one true positive within the retrieval set of size $k$, according to the aforementioned definitions of ground truth.

\subsection{Image Pair Retrieval Evaluation}\label{sec:image-pair-retrieval-eval}

We aim to perform evaluation on datasets Tanks and Temples (T\&T)~\citep{Knapitsch2017} and ScanNet-GSReg~\citep{dai2017scannet,Chang:2024} based on SfM-reconstructions~\citep{schoenberger2016sfm}, which yield camera poses and intrinsics for further tasks.
KITTI-odometry (KITTI)~\citep{Geiger2012CVPRkitti} as our third dataset is used with its own ground truth camera poses and intrinsics. 
All three datasets contain either scenes (T\&T, ScanNet-GSReg) or sequences (KITTI), which we also denote as scenes throughout this work. 
A scene is split into subsets $A$ and $B$ with both sets including subsets of images showing the same scene area. 
We evaluate SOTA methods regarding their ability to match image pairs that cover such areas. 
The top-$k$ image pair matches are utilized as the retrieval set in order for tasks such as RGB-D registration to yield a sufficient baseline for matching images. 
Although in all three datasets images were acquired as sequences, we perform a brute-force matching without prior information.

\paragraph{Metrics.}
As described in Sec.~\ref{sec:preliminaries-vpr-eval}, \textit{Recall@k} is the standard metric in VPR, which we also include in our task of image pair matching, yielding information if there is at least one true positive match within the retrieval set. 
However, information about the \textit{quantity} of true positives within top-$k$ pairs is relevant especially for image-driven scene registration. 
This holds true if the number of true positives is too low, so that even robust methods like RANSAC-based transformation estimation might fail in a registration scenario. 
We therefore employ the \textit{Precision@k} (P@k) metric to gather information about the ratio of true positives to false positives within the retrieval set. 
As a third metric, we utilize \textit{mean-Average-Precision@k} (mAP@k), which exploits information about the ranking of true positives within a retrieval set.
We therefore aim to gather information about how to set $k$.\\
All three metrics are averaged across all scenes per dataset. By doing so, we follow the usual practice in VPR evaluation, where R@k is averaged across all queries, i.e. all retrieval sets per dataset. Also, as common in VPR literature, the retrieval set sizes are $k\in\{1, 5, 10\}$.

\paragraph{Matching.}
For brute-force matching of images $I_i^A \in \mathcal{I}_A$ and $I_j^B \in \mathcal{I}_B$ 
we first compute a global descriptor of dimension $D$ for every image using the employed VPR method $\psi : I \to \mathbb{R}^D$. We then define the descriptor sets as
\begin{equation}
    \mathcal{D}_X 
    = \{ \mathbf{d}_k^X = \psi(I_k^X) \mid I_k^X \in \mathcal{I}_X \},
    \quad X \in \{A,B\}.
\end{equation} 
In the second step, the top-$k$ descriptor pairs are obtained by computing the cosine
similarity across all possible pairs and ranking them in descending order:
\begin{equation}
    \mathcal{P}_k 
    = \operatorname*{arg\,top-k}_{i,j}
      \frac{(\mathbf{d}_i^A)^\top \mathbf{d}_j^B}
           {\lVert \mathbf{d}_i^A\rVert_2 \,\lVert \mathbf{d}_j^B\rVert_2},
\end{equation}
where $\mathcal{P}_k$ contains the indices $(i,j)$ of the top-$k$ pairs, sorted by decreasing cosine similarity.
\paragraph{Ground truth definition.}
Based on SfM~\citep{schoenberger2016sfm} camera poses for T\&T and SN-GS and ground truth poses from KITTI, the first criterion to be met is the view direction angle of two images. 
While the threshold is set to $40^{\circ}$ in VPR literature, we found that specifically for 
object-centric scenarios, a widening of this angle leads to higher completeness and less rejection of valid correspondences in the evaluation process.
We therefore set the threshold to $\tau_{view} = 75^{\circ}$. 
Let $R_i^A, R_j^B \in SO(3)$ denote the world-to-camera rotation matrices of images $A$ and $B$, respectively.
Their viewing directions in world coordinates are defined as
\begin{equation}
    \mathbf{d}_A = (R_i^A)^\top \begin{bmatrix}0 \\ 0 \\ -1\end{bmatrix}, 
    \qquad
    \mathbf{d}_B = (R_j^B)^\top \begin{bmatrix}0 \\ 0 \\ -1\end{bmatrix}.
\end{equation}
The angle between both viewing directions is then given by
\begin{equation}
    \phi_{\text{view}} 
    = \arccos\!\left(
        \frac{\mathbf{d}_A^\top \mathbf{d}_B}
             {\lVert \mathbf{d}_A\rVert_2 \,\lVert \mathbf{d}_B\rVert_2}
    \right),
\end{equation}
and the first criterion is fulfilled if
\begin{equation}
    \phi_{\text{view}} \le \tau_{view}.
\end{equation}
Opposed to the benchmark dataset used for VPR~\citep{sunderhauf2013we_Nordland, vpr_repetetive_torii2013, tokio247_torii2015, Warburg2020_Mapillary_StreetLevel_CVPR}, our camera poses from SfM reconstructions are only defined up to a scale w.r.t.\ their position. 
Therefore, we cannot utilize a reliable distance measure as common in VPR methods and pointed out in Sec.~\ref{sec:preliminaries-vpr-eval}. However, in computer vision tasks such as the estimation of homography, fundamental or essential matrix, and 3D transformations, it is common to rely on matched keypoints. 
Therefore, we define a second criterion as follows: We extract and match SIFT keypoints~\citep{Lowe:2004:Sift} from image pairs and, given intrinsic camera parameters, estimate the essential matrix $E$ via RANSAC-based 5-point algorithm~\citep{Nister_5Point}. 
Utilizing SVD, $E$ is decomposed into rotation matrix $R^{A\to B}_E\in SO3$ and unit translation vector $||t_i||_2$. 
For better readability, mapping $A\to B$ is left out in the following. 
We check if $R_E$ approximates the rotation matrix $R_{SfM}$ of the relative pose of both images using the geodesic distance of both rotation matrices:
\begin{equation}
    d_R\!\left(R_E, R_{\mathrm{SfM}}\right)
    = \arccos\!\left(
        \frac{{tr}\!\big(R_E R_{\mathrm{SfM}}^\top\big) - 1}{2}
    \right),
\end{equation}
and set a threshold of $\tau_{dev}=10^{\circ}$ maximum angular deviation, so that
\begin{equation}
    \frac{d_R\cdot 180 ^{\circ}}\pi < \tau_{dev}.
\end{equation}
To add further robustness, an inlier threshold of \mbox{$\tau_{in}=0.25$} is set for the estimation of $E$. 
For the purpose of comparability, this second criterion is also applied for the KITTI scenes, even though the view position is given in absolute units (meters) per image and therefore could be compared analog to VPR evaluation, i.e. through a distance threshold.\\
We also account for the runtime without applying typical re-ranking, as this step often requires several seconds and would exceed our goal of fast initial image pair retrieval~\citep{Ali-bey_2023_WACV_MixVPR,Izquierdo_CVPR_2024_SALAD,hausler2021patchnetvlad}.

\section{Experiments}\label{sec:experiments}
In this section, we describe the experimental setup with the datasets, methods, and the used hardware.
\subsection{Datasets}\label{sec:datasets}
In order to evaluate the described methods, we customized T\&T~\citep{Knapitsch2017} and KITTI ~\citep{Geiger2012CVPRkitti} scenes by dividing them into two disjoint subsets $A$ and $B$ with overlapping scene content in both subsets as already pointed out in Sec.~\ref{sec:image-pair-retrieval-eval}. 
From T\&T, we utilized the \textit{Caterpillar} \textit{Barn}, \textit{Train}, \textit{Truck}, \textit{Palace}, \textit{Playground} and \textit{Lighthouse} scenes as the object-centric scenarios. With object centric trajectories, we want to focus on ambiguous texture and repeating objects in disjoint scene regions, which is usually refered to as perceptual aliasing. 
From the KITTI dataset, we utilized sequences \textit{00}, \textit{02}, \textit{03}, \textit{05}, \textit{06}, \textit{07}, \textit{08}, \textit{09}, and \textit{10}. 
We left out sequences with a significant amount of moving objects, because of our focus on largely static scenes within this research. 
It should be noted that for both T\&T and KITTI, one or more subsets are extracted per scene and split into $A$ and $B$ per subset. $A$ and $B$ contain between 30 and 60 images. KITTI represents an outdoor scenario with the focus on larger (sub-)urban environments, similar to large-scale benchmark datasets used for VPR training and evaluation, but with an emphasis on autonomous driving.\\
For indoor scenarios, we utilize the ScanNet-GSReg (SN-GS) dataset from~\cite{Chang:2024}, which is an already prepared dataset for 3DGS registration, based on the ScanNet dataset~\citep{dai2017scannet}. 
We utilized scenes \textit{0000\_01}, \textit{0009\_00}, \textit{0018\_00}, \textit{0050\_00}, \textit{0100\_02}, \textit{0111\_01}, \textit{0170\_02}, \textit{0218\_01}, \textit{0222\_01}, \textit{0309\_01}, \textit{0328\_00}, \textit{0369\_01}, \textit{0420\_00}, \textit{0455\_00}, \textit{0541\_00}, \textit{0568\_00}, \textit{0588\_00}, \textit{0591\_01}, \textit{0629\_00}, \textit{0630\_01}, \textit{0666\_00}, \textit{0667\_00}, \textit{0682\_00}, \textit{0696\_00}, \textit{0701\_00} and \textit{0703\_00} in our experiments. The scenes were chosen randomly, as all scenes are indoor with similar acquisition configurations. It is important to note that MegaLoc~\citep{Berton_2025_MegaLoc}, which is used in our evaluation, was trained on the original ScanNet dataset and therefore is \textit{heavily biased} regarding the aforementioned scenes. Nevertheless, we included the results of this method on this dataset for the sake of completeness. Another important aspect of this dataset is that some scenes contain a small number of identical images in $A$ and $B$. We did not modify these scenes, in order for future research to be applied on the SN-GS dataset.\\
In total, we use 16 scenes from T\&T, 26 scenes from SN-GS, and 26 scenes from KITTI.
\subsection{Methods}
We compare a wide range of SOTA methods based on the datasets from Sec.~\ref{sec:datasets} and metrics from Sec.~\ref{sec:image-pair-retrieval-eval}. The methods include PatchNetVLAD~\citep{hausler2021patchnetvlad}, CosPlace~\citep{Berton_CVPR_2022_CosPlace}, EigenPlaces~\citep{Berton_2023_EigenPlaces}, MixVPR~\citep{Ali-bey_2023_WACV_MixVPR}, AnyLoc~\citep{keetha2023anyloc}, SALAD~\citep{Izquierdo_CVPR_2024_SALAD}, and MegaLoc~\citep{Berton_2025_MegaLoc}. 
The methods CosPlace, EigenPlaces, SALAD, and MixVPR provide different backbone or descriptor configurations. For our evaluation, different configurations of those methods are used as listed in Table~\ref{tab:vpr_configs}.
\begin{table}
    \centering
    \small
    \setlength{\tabcolsep}{3pt}
    \begin{tabular}{c | c c}
    Method              & Backbone      & Descriptor size \\
    \hline
    CosPlace512         & ResNet18      & 512 \\
    CosPlace2048        & ResNet101     & 2048 \\
    EigenPlaces512      & ResNet18      & 512 \\
    EigenPlaces2048     & ResNet101     & 2048 \\
    MixVPR512           & ResNet50      & 512 \\
    MixVPR4096          & ResNet50      & 4096 \\
    SALAD2048           & DINOv2-b14    & 2048 \\
    SALAD8192           & DINOv2-b14    & 8192 
    \end{tabular}
    \caption{Different backbone and descriptor size configurations used for the methods CosPlace~\citep{Berton_CVPR_2022_CosPlace}, EigenPlaces~\citep{Berton_2023_EigenPlaces}, MixVPR~\citep{Ali-bey_2023_WACV_MixVPR}, and SALAD~\citep{Izquierdo_CVPR_2024_SALAD}.}
    \label{tab:vpr_configs}
\end{table}
PatchNetVLAD as an older method is included, because it is close to the original NetVLAD-based method and further includes local reranking. Regarding AnyLoc, we did not use an initialization with specific cluster centers, as such cluster centers are biased towards certain domains they have been optimized to. AnyLoc, SALAD and MegaLoc utilize ViT backbones (DINOv2), whereas MixVPR, CosPlace, EigenPlaces and PatchNetVLAD utilize CNN backbones (ResNet).

\subsection{Hardware}
For all methods, the experiments were conducted on a PC with an Nvidia RTX 3090 GPU and an Intel i9 10850 CPU with 32 Gb RAM.

\section{Results}\label{sec:results}
\begin{figure*}[!ht]
  \centering
  \setlength{\tabcolsep}{1pt}
  \renewcommand{\arraystretch}{1.0}
  \begin{tabular}{ccc}
    a) PatchNetVLAD & b) CosPlace512 & c) SALAD8192 \\[2pt]
    \includegraphics[width=\topfivewidth]{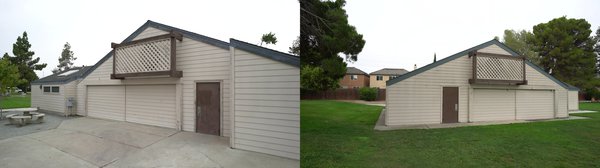} &
    \includegraphics[width=\topfivewidth]{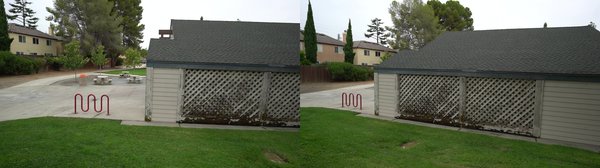} &
    \includegraphics[width=\topfivewidth]{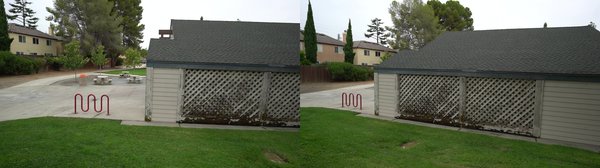} \\
    \includegraphics[width=\topfivewidth]{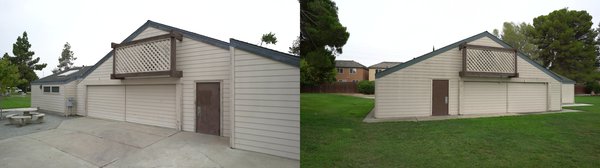} &
    \includegraphics[width=\topfivewidth]{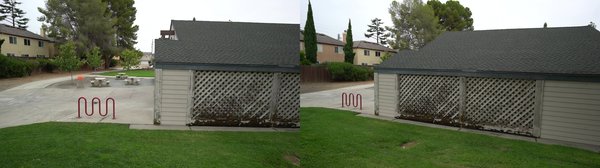} &
    \includegraphics[width=\topfivewidth]{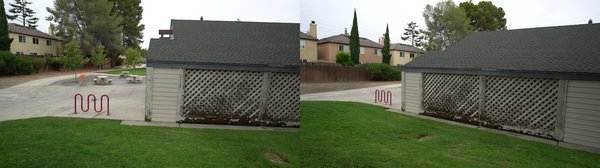} \\
    \includegraphics[width=\topfivewidth]{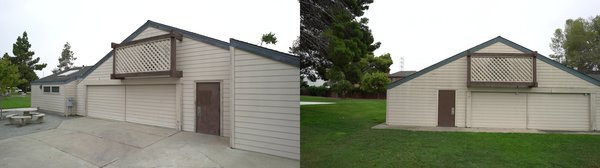} &
    \includegraphics[width=\topfivewidth]{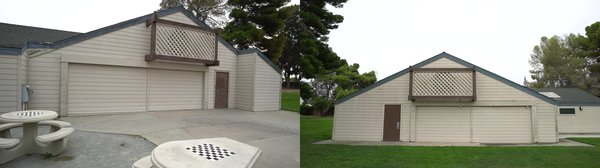} &
    \includegraphics[width=\topfivewidth]{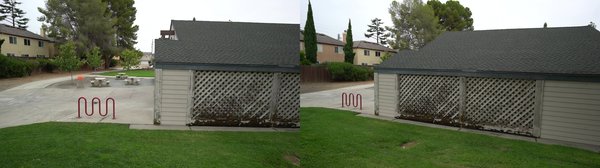} \\
    \includegraphics[width=\topfivewidth]{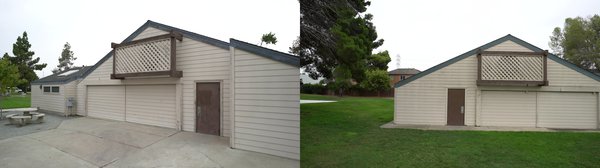} &
    \includegraphics[width=\topfivewidth]{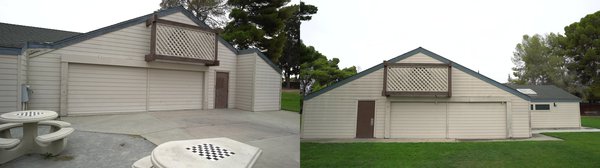} &
    \includegraphics[width=\topfivewidth]{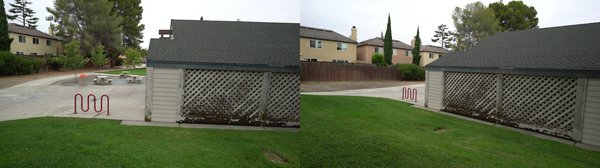} \\
    \includegraphics[width=\topfivewidth]{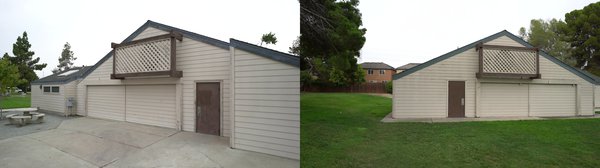} &
    \includegraphics[width=\topfivewidth]{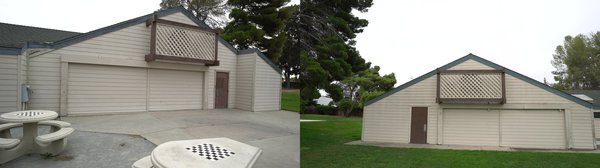} &
    \includegraphics[width=\topfivewidth]{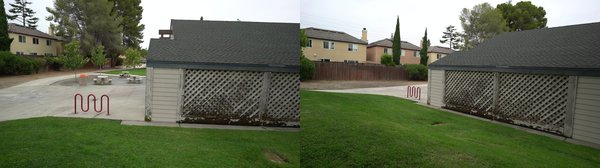} \\
    [4pt]
    d) EigenPlaces512 & e) MegaLoc & f) MixVPR4096 \\[2pt]
    \includegraphics[width=\topfivewidth]{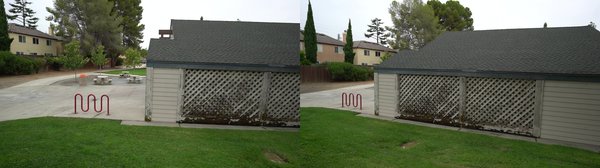} &
    \includegraphics[width=\topfivewidth]{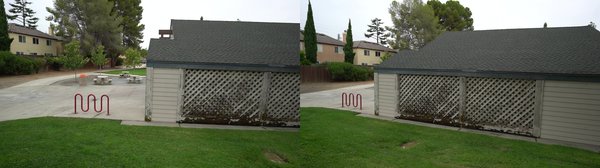} &
    \includegraphics[width=\topfivewidth]{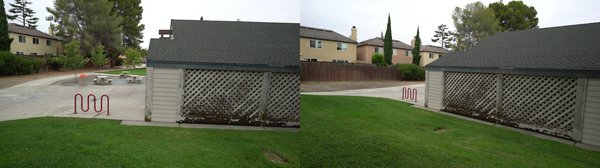} \\
    \includegraphics[width=\topfivewidth]{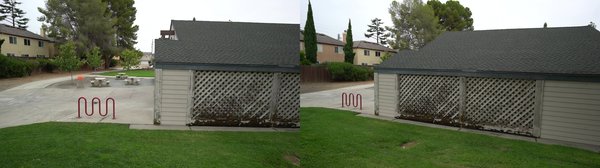} &
    \includegraphics[width=\topfivewidth]{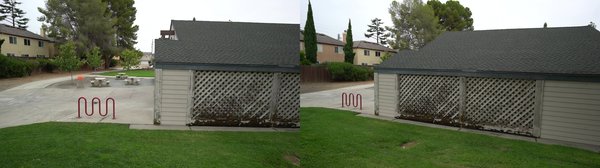} &
    \includegraphics[width=\topfivewidth]{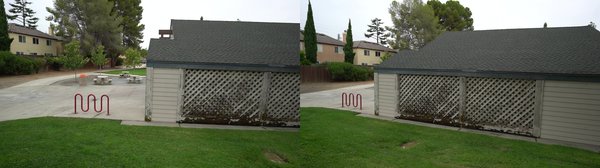} \\
    \includegraphics[width=\topfivewidth]{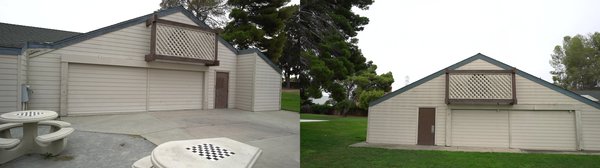} &
    \includegraphics[width=\topfivewidth]{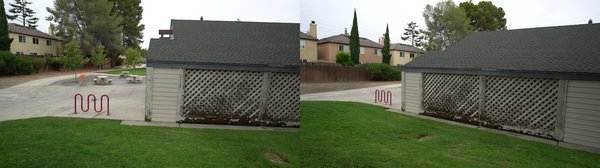} &
    \includegraphics[width=\topfivewidth]{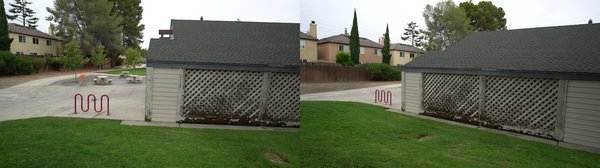} \\
    \includegraphics[width=\topfivewidth]{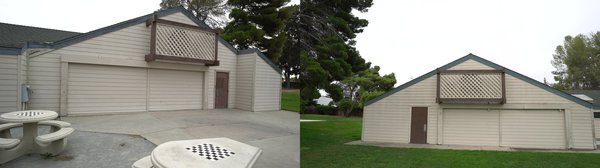} &
    \includegraphics[width=\topfivewidth]{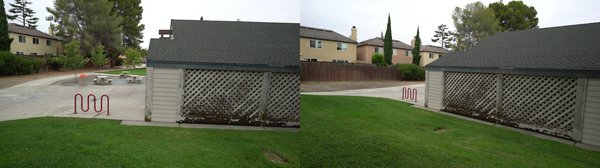} &
    \includegraphics[width=\topfivewidth]{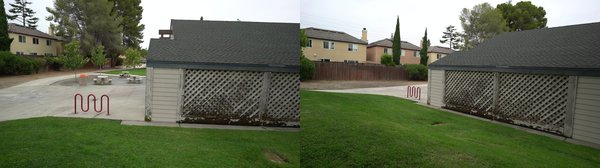} \\
    \includegraphics[width=\topfivewidth]{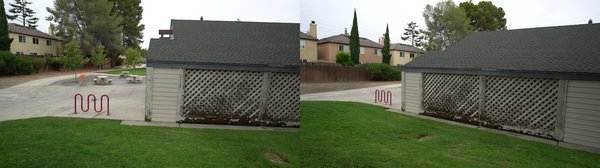} &
    \includegraphics[width=\topfivewidth]{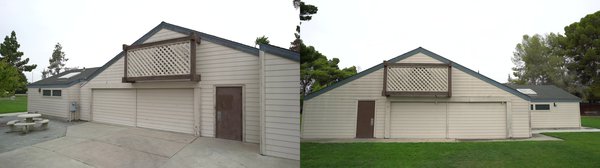} &
    \includegraphics[width=\topfivewidth]{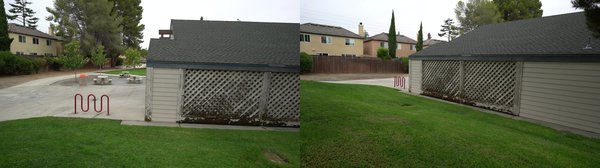} \\
  \end{tabular}
  \caption{Qualitative top-5 retrieval results for different VPR methods on the T\&T scene \textit{Barn}. This scene is especially prone to perceptual aliasing, as indicated by the brown door, which is at different positions on both sides and the color of the floor in front of the house. Overlap areas are the back of the house, which is completely correctly retrieved by SALAD8192 and MixVPR4096. Columns correspond to methods (a--f) and, within each column, images are shown from rank $k=1$ (top) to $k=5$ (bottom).}
  \label{fig:tnt_barn01_top5_all}
\end{figure*}

\begin{figure*}[!t]
  \centering
  \setlength{\tabcolsep}{1pt}
  \renewcommand{\arraystretch}{1.0}
  \begin{tabular}{c}
    a) CosPlace512 \\[2pt]
    \includegraphics[width=\kittifivewidth]{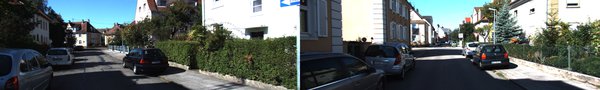} \\
    \includegraphics[width=\kittifivewidth]{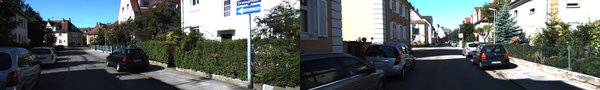} \\
    \includegraphics[width=\kittifivewidth]{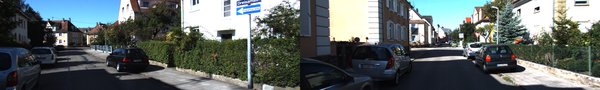} \\
    \includegraphics[width=\kittifivewidth]{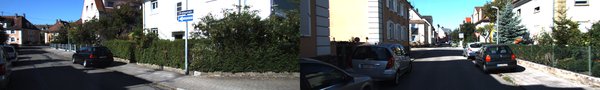} \\
    \includegraphics[width=\kittifivewidth]{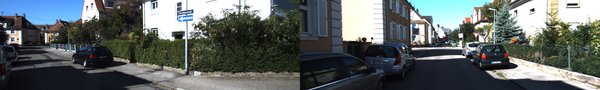} \\
    [6pt]
    b) MegaLoc \\[2pt]
    \includegraphics[width=\kittifivewidth]{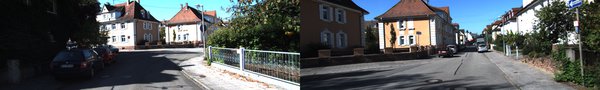} \\
    \includegraphics[width=\kittifivewidth]{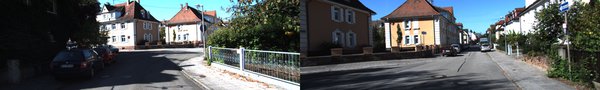} \\
    \includegraphics[width=\kittifivewidth]{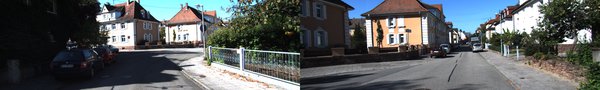} \\
    \includegraphics[width=\kittifivewidth]{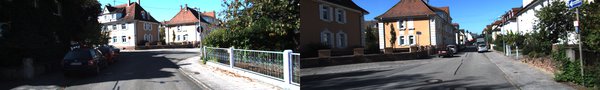} \\
    \includegraphics[width=\kittifivewidth]{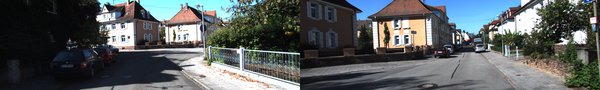} \\
  \end{tabular}
  \caption{Qualitative Top-5 retrieval results from CosPlace512 (a) and MegaLoc (b) for KITTI scene 07\_02. The challenge here is that some images were left out between $A$ and $B$. This is especially important for tasks such as SLAM relocalization caused from missing frames. CosPlace~\citep{Berton_CVPR_2022_CosPlace} completely fails here, likely because of perceptual aliasing, indicated by two different black cars in $A$ and $B$ and other similarities, whereas MegaLoc~\citep{Berton_2025_MegaLoc} yields only true positives. The cue for correct results is the houses in the center of the $A$ images, which appear towards the left border of the $B$ images. Images are shown from $k=1$ (top) to $k=5$ (bottom) per method.}
  \label{fig:kitti_0702_cosplace_megaloc}
\end{figure*}

\paragraph{Qualitative Results}
We show qualitative results in Figure~\ref{fig:tnt_barn01_top5_all} for top-5 retrievals of six methods for the T\&T Barn scene and top-5 retrievals for two selected methods in Figure~\ref{fig:kitti_0702_cosplace_megaloc} for a scene \textit{07\_02} of the KITTI dataset~\citep{Geiger2012CVPRkitti}. Both are selected for certain perceptual aliasing challenges and wide-baseline image matching to show the performance of current SOTA VPR methods regarding image pair retrieval.

\paragraph{Quantitative Results}
The quantitative results of the image pair retrievals across all datasets are shown in Table~\ref{tab:vpr_results}. Metrics P@k, R@k, and mAP@k are utilized as described in Sec.~\ref{sec:image-pair-retrieval-eval} with $k\in \{1,5,10\}$ for P@k and R@k. For mAP@k, only $k\in\{5,10\}$ are used, as mAP@1 yields the same results as R@1. P@10 and R@10 are represented graphically in Fig.~\ref{fig:vpr_p10_r10_overview}.
\begin{table*}[!t]
    \centering
    \small
    \setlength{\tabcolsep}{4pt}
    \begin{tabular}{c c | c c c  c c c  c c   | c c}
    Dataset & Method    & P@1   & P@5   & P@10  & R@1   & R@5   & R@10  & mAP@5 & mAP@10&$\mu_t$&$\sigma_t$ \\
    \hline
     & PatchNetVLAD         & 25.00 & 21.25 & 17.50 & 25.00 & 37.50 & 43.75 & 25.10 & 26.27 & 1.102 & 0.105 \\
     & CosPlace512      & 56.25 & 43.75 & 36.88 & 56.25 & 56.25 & 62.50 & 55.62 & 52.69 & 0.133 & 0.014 \\
     & CosPlace2048     & 56.25 & 40.00 & 32.50 & 56.25 & 56.25 & 62.50 & 53.33 & 51.54 & 0.645 & 0.070 \\
     & EigenPlaces512   & 56.25 & 47.50 & 36.88 & 56.25 & 56.25 & 62.50 & 54.37 & 54.35 & 0.134 & 0.016 \\
     & EigenPlaces2048  & 56.25 & 45.00 & 35.62 & 56.25 & 62.50 & 62.50 & 57.19 & 53.67 & 0.649 & 0.070 \\
    T\&T & MixVPR512    & 18.75 & 6.25  & 5.00  & 18.75 & 25.00 & 31.25 & 21.88 & 21.56 & 0.130 & 0.012 \\
     & MixVPR4096       & 25.00 & 13.75 & 8.12  & 25.00 & 25.00 & 31.25 & 23.44 & 23.28 & 0.133 & 0.012 \\
     & AnyLoc           & 31.25 & 21.25 & 16.25 & 31.25 & 43.75 & 50.00 & 32.84 & 31.04 & 6.116 & 3.771 \\
     & SALAD2048        & 56.25 & 46.25 & 34.38 & 56.25 & 56.25 & 56.25 & 55.03 & 53.70 & 2.740 & 0.305 \\
     & SALAD8192        & 50.00 & 38.75 & 29.38 & 50.00 & 50.00 & 56.25 & 47.25 & 46.78 & 2.755 & 0.307 \\
     & MegaLoc          & 56.25 & 47.50 & 35.62 & 56.25 & 56.25 & 56.25 & 53.46 & 51.72 & 2.799 & 0.316 \\
    \hline
     & PatchNetVLAD         & 92.31 & 93.08 & 93.46 & 92.31 & 96.15 & 96.15 & 93.21 & 93.47 & 2.541 & 0.187 \\
     & CosPlace512      & 100.00& 98.46 & 99.23 & 100.00& 100.00& 100.00& 98.81 & 98.92 & 0.375 & 0.028 \\
     & CosPlace2048     & 100.00& 99.23 & 99.23 & 100.00& 100.00& 100.00& 99.81 & 99.55 & 1.817 & 0.126 \\
     & EigenPlaces512   & 100.00& 99.23 & 99.23 & 100.00& 100.00& 100.00& 99.57 & 99.44 & 0.372 & 0.025 \\
     & EigenPlaces2048  & 100.00& 98.46 & 99.23 & 100.00& 100.00& 100.00& 98.85 & 98.95 & 1.825 & 0.124 \\
   SN-GS & MixVPR512    & 96.15 & 94.62 & 93.85 & 96.15 & 96.15 & 100.00& 95.40 & 96.06 & 0.301 & 0.022 \\
     & MixVPR4096       & 92.31 & 91.54 & 93.08 & 92.31 & 100.00& 100.00& 94.04 & 93.59 & 0.308 & 0.023 \\
     & AnyLoc           & 92.31 & 94.62 & 93.85 & 92.31 & 96.15 & 100.00& 94.36 & 94.73 & 12.501& 2.821 \\
     & SALAD2048        & 100.00& 100.00& 99.62 & 100.00& 100.00& 100.00& 100.00& 100.00& 9.458 & 0.669 \\
     & SALAD8192        & 100.00& 99.23 & 99.62 & 100.00& 100.00& 100.00& 99.57 & 99.53 & 9.560 & 0.683 \\
      & MegaLoc         & 100.00& 100.00& 100.00& 100.00& 100.00& 100.00& 100.00& 100.00& 9.642 & 0.694 \\
    \hline
     & PatchNetVLAD         & 34.62 & 24.62 & 21.54 & 34.62 & 38.46 & 38.46 & 34.36 & 31.47 & 1.683 & 0.293 \\
     & CosPlace512      & 34.62 & 35.38 & 33.46 & 34.62 & 38.46 & 38.46 & 36.54 & 35.82 & 0.120 & 0.045 \\
     & CosPlace2048     & 34.62 & 35.38 & 32.31 & 34.62 & 38.46 & 38.46 & 36.54 & 36.36 & 0.551 & 0.094 \\
     & EigenPlaces512   & 34.62 & 34.62 & 32.31 & 34.62 & 34.62 & 34.62 & 34.62 & 34.18 & 0.112 & 0.019 \\
     & EigenPlaces2048  & 34.62 & 34.62 & 32.31 & 34.62 & 34.62 & 38.46 & 34.62 & 34.91 & 0.546 & 0.092 \\
     KITTI & MixVPR512  & 19.23 & 19.23 & 16.92 & 19.23 & 26.92 & 30.77 & 23.25 & 23.18 & 0.198 & 0.035 \\
     & MixVPR4096       & 30.77 & 22.31 & 20.77 & 30.77 & 38.46 & 38.46 & 32.97 & 31.76 & 0.204 & 0.036 \\
     & AnyLoc           & 23.08 & 23.08 & 21.92 & 23.08 & 26.92 & 30.77 & 24.17 & 24.10 & 8.516 & 3.163 \\
     & SALAD2048        & 34.62 & 35.38 & 33.08 & 34.62 & 38.46 & 38.46 & 36.54 & 36.54 & 1.799 & 0.311 \\
     & SALAD8192        & 34.62 & 34.62 & 33.08 & 34.62 & 34.62 & 34.62 & 34.62 & 34.55 & 1.813 & 0.316 \\
     & MegaLoc          & 34.62 & 35.38 & 33.08 & 34.62 & 38.46 & 38.46 & 36.54 & 36.47 & 1.848 & 0.319 \\
\end{tabular}
    \caption{Image pair retrieval across different datasets, consisting of multiple scenes (quantity in braces) for each dataset: T\&T (16), SN-GS (26), and KITTI (26). Each dataset is evaluated on seven VPR methods, split into eleven configurations. Metrics are calculated for the $k$ best ranked image pairs, leading to Precision@k (P@k), Recall@k (R@k), and mean-Average-Precision@k (mAP@k), all given in percent. Average time in seconds for all retrievals
    per scene is included as $\mu_t$ with $\sigma_t$ as the standard deviation, respectively.}
    \label{tab:vpr_results}
\end{table*}
\begin{figure*}[!t]
  \centering
  \setlength{\tabcolsep}{2pt}
  \renewcommand{\arraystretch}{1.0}
  \begin{tabular}{cc}
    \includegraphics[width=0.38\textwidth]{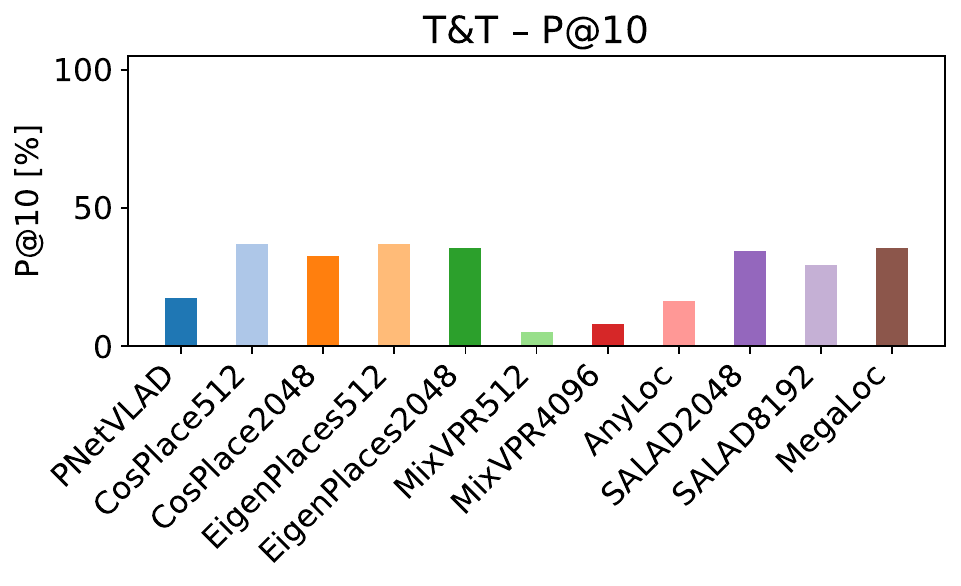} &
    \includegraphics[width=0.38\textwidth]{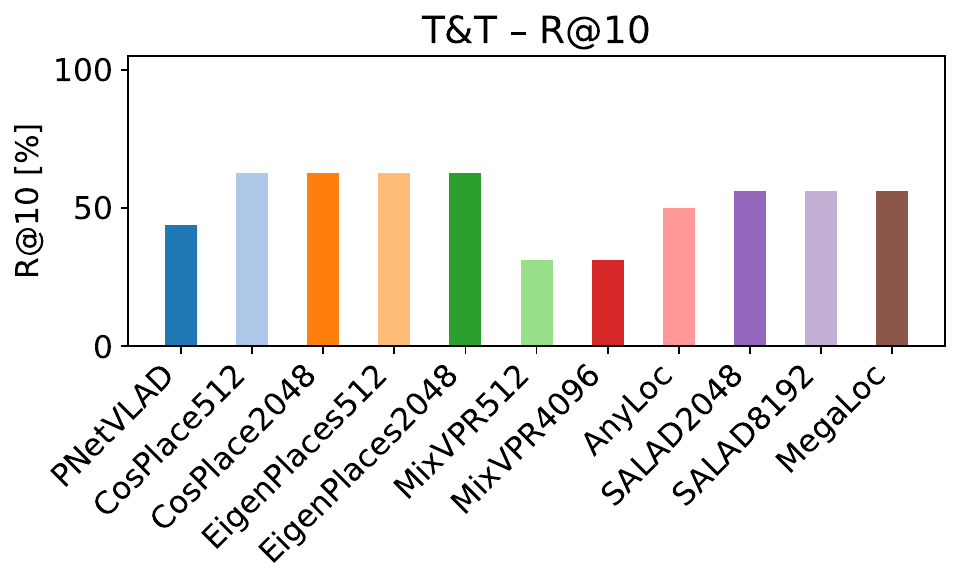} \\
    \includegraphics[width=0.38\textwidth]{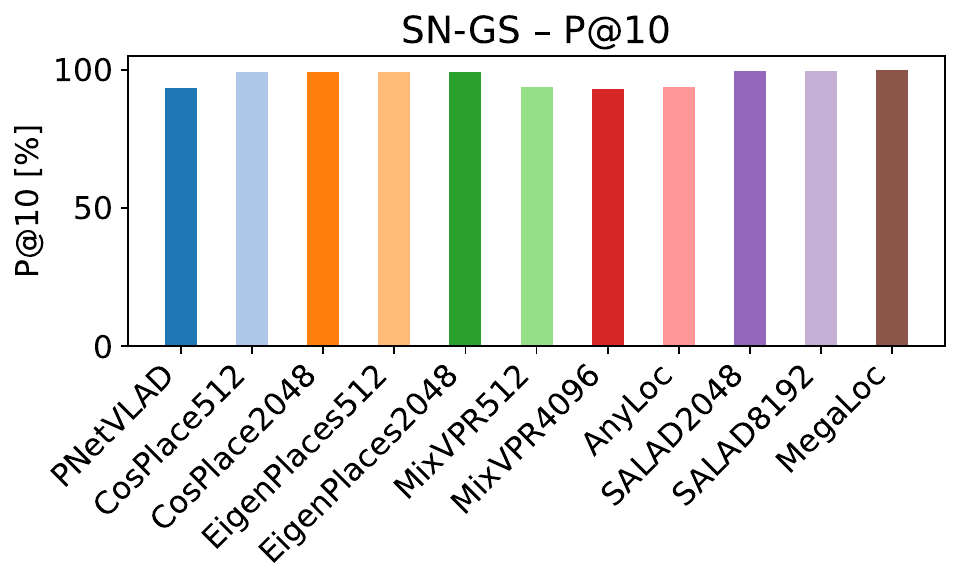} &
    \includegraphics[width=0.38\textwidth]{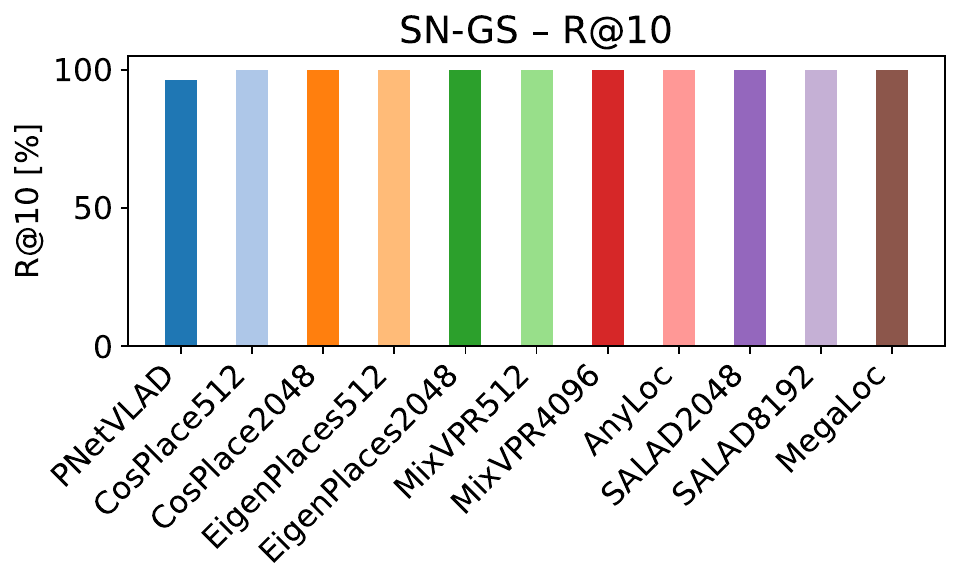} \\
    \includegraphics[width=0.38\textwidth]{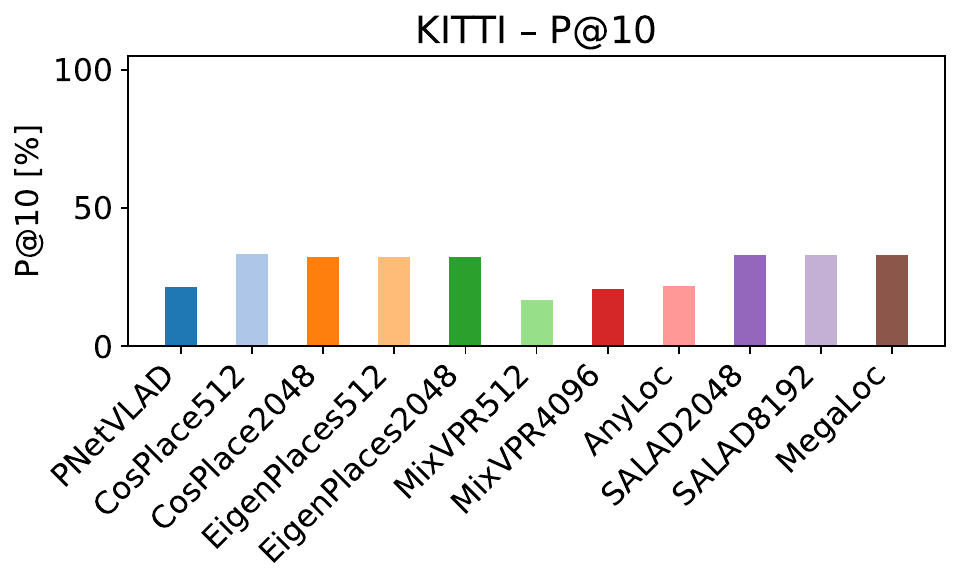} &
    \includegraphics[width=0.38\textwidth]{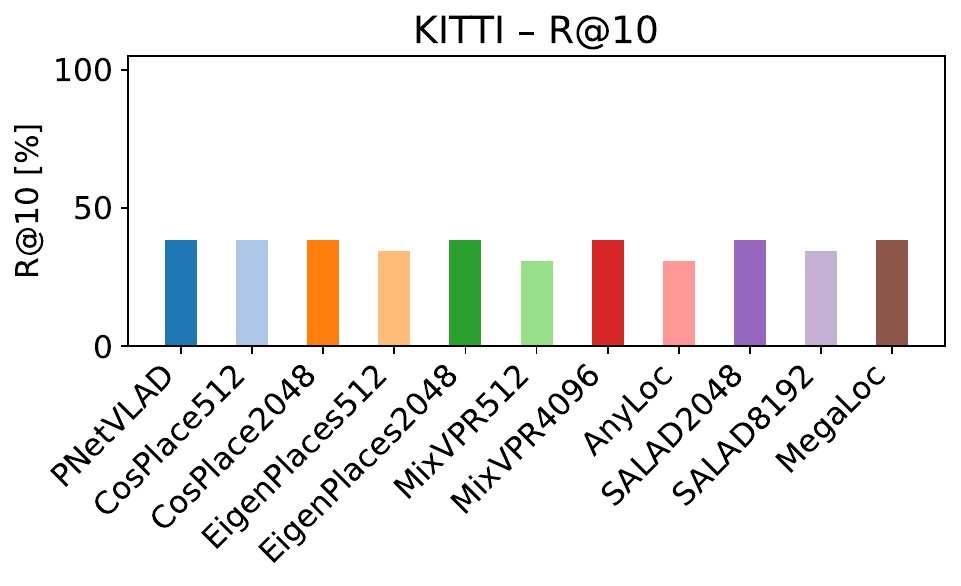} \\
  \end{tabular}
  \caption{Overview of P@10 (left column) and R@10 (right column) for all VPR methods on T\&T (top row), SN-GS (middle row) and KITTI (bottom row).}
  \label{fig:vpr_p10_r10_overview}
\end{figure*}

\section{Discussion}\label{sec:discussion}
The results in Table~\ref{tab:vpr_results} show different retrieval performances per dataset. Due to the matters regarding the SN-GS as mentioned in Sec.~\ref{sec:datasets}, the high performances are easily explainable. It is to be expected that identical images are not only found (R@k), but also ranked at the top (mAP@k). Also as mentioned already in Sec.~\ref{sec:datasets}, MegaLoc~\citep{Berton_2025_MegaLoc} as ViT-based method was trained on ScanNet among others, which probably leads to overly optimistic retrievals across all scenes in the dataset. 
Cross-domain generalization for this method is therefore better assessed in T\&T and KITTI datasets, where it outperforms the other methods in most cases. 
Other ViT-based methods are SALAD~\citep{Izquierdo_CVPR_2024_SALAD} and AnyLoc~\citep{keetha2023anyloc}, whereby AnyLoc performs poorest of this group. As stated in~\citep{Izquierdo_CVPR_2024_SALAD}, a fine-tuning of the general purpose ViT-model (DINOv2) can improve performance, which was done for the SALAD method as opposed to AnyLoc. The CNN-based methods mostly perform a bit worse than the ViT methods. EigenPlaces~\citep{Berton_2023_EigenPlaces} as a CNN-based method in both configurations is close to the ViT-based method, and outperforms AnyLoc across all metrics.\\
Across all methods, there is no significant drop between mAP@5 and mAP@10. This induces, that a small $k$ can be sufficient if computation times for following tasks (e.g., transformation estimation) are critical. For robustness and offline use, a higher $k$ is to be preferred, if both P@k and mAP@k yield sufficient numbers. For P@k, usually 20~\% are the lower bound for robust methods like RANSAC to obtain enough inliers. It is also worth noting that the descriptor size and backbone depth in CosPlace~\citep{Berton_CVPR_2022_CosPlace} and EigenPlaces~\citep{Berton_2023_EigenPlaces} only have a small impact. In some metrics (e.g., in T\&T), the smaller models with smaller descriptor sizes even perform better in P@k, meaning over all retrieved pairs, the number of true positives is higher. The parameter $p$ of the GeM-pooling that interpolates between average- and maximum-pooling is between 2.6 (CosPlace2048) and 3.0 (Eigenplaces512), the other two being at around 2.9. This pooling parameter has direct impact on the descriptor. However, it is unclear if it correlates to model complexity and descriptor size. For SALAD~\citep{Izquierdo_CVPR_2024_SALAD}, small descriptor sizes seem to perform better across almost all metrics and datasets. For MixVPR~\citep{Ali-bey_2023_WACV_MixVPR}, the opposite is the case. The size of the descriptor on its own is not a clear indicator of retrieval performance. We therefore opt for experimentation and choose the best working configuration for a given task.\\
To have a more detailed view of the retrieved image pairs, Fig.~\ref{fig:tnt_barn01_top5_all} shows the top-5 retrievals for six methods. This scene was chosen because it yields the challenge of perceptual aliasing. PatchNetVLAD~\citep{hausler2021patchnetvlad} obtains false positives only, even though the local reranking would induce a taming of perceptual aliasing in the form of similar content at different positions, e.g., as is the case for the brown door in the images. Even MegaLoc~\citep{Berton_2025_MegaLoc} includes a false positive in the top-5. SALAD~\citep{Izquierdo_CVPR_2024_SALAD}, and MixVPR~\citep{Ali-bey_2023_WACV_MixVPR} exceed in this case. MixVPR, however, also performs worst on T\&T in general. In other cases of perceptual aliasing, where similar objects also appear at the similar position in the image, MixVPR fails. The domain gap between the training data and the test data might as well be too large for MixVPR to generalize well to such object-centric scenes. Fig.\ref{fig:kitti_0702_cosplace_megaloc} shows two methods for KITTI scene with the additional challenge of missing frames, besides the perceptual aliasing. We chose CosPlace~\citep{Berton_CVPR_2022_CosPlace}, which completely fails for the top-5, and MegaLoc~\citep{Berton_2025_MegaLoc}, which exceeds in this example. It is notable that MegaLoc appears to be exceptionally robust against perceptual aliasing, which is mostly represented by cars and vegetation, whereas for CosPlace~\citep{Berton_CVPR_2022_CosPlace} it is clearly visible that it fails for such aliasing.\\
Finally, the runtime performance indicates clearly that CNN-based methods outperform ViT-based methods by a large margin. With the exception of PatchNetVLAD~\citep{hausler2021patchnetvlad}, all CNN-based methods perform below one second, whereas the ViT-based methods take more than one or multiple seconds on average. This is important for real-time tasks. Both configurations of EigenPlaces~\citep{Berton_2023_EigenPlaces} seem to yield the best balance of runtime and retrieval performance.
\\
Summarized, our experiments suggest the following practical insights:
\begin{itemize}
    \item \textbf{Object-centric scenes (T\&T):} Scenes with strong perceptual aliasing (e.g., repeated structures such as wheels or façades) remain challenging for all methods. ViT-based models such as SALAD and MegaLoc often produce more robust top-$k$ lists in the most ambiguous cases (cf.\ Fig.~2), while strong CNN-based methods like EigenPlaces achieve comparable average P@k and R@k at substantially lower runtime.
    \item \textbf{Indoor scenes (SN-GS):} All methods perform extremely well, which we attribute to the high overlap between the disjoint sets A and B and, in some scenes, even a small number of identical images across both sets.
    \item \textbf{Outdoor autonomous navigation (KITTI):} For typical SLAM and odometry scenarios, CNN-based methods such as EigenPlaces offer a favorable trade-off between retrieval quality and runtime. The performance gap in P@k and R@k to the best-performing ViT-based methods is only minor, while the runtime advantage is substantial.
\end{itemize}
\section{Conclusion}\label{sec:conclusion}

In this work, we presented an extensive evaluation of state-of-the-art VPR methods for the task of image pair retrieval. Our evaluation pipeline is specifically designed for photogrammetrically reconstructed scenes, and our metrics are tailored to typical 3D vision and robotics tasks. We showed that VPR methods that were especially trained for out-of-domain performance (e.g., MegaLoc) can be used as strong off-the-shelf front-ends for the initial matching of images from two disjoint datasets. This is particularly relevant for RGB-D registration, 3D Gaussian Splatting registration, relocalization, loop-closure detection, and boosting SfM correspondence search. In real-time scenarios, CNN-based methods appear preferable due to their favourable trade-off between retrieval quality and runtime, whereas ViT-based methods achieve higher retrieval performance in highly challenging scenes with strong perceptual aliasing and missing frames in image sequences.\\
By design, this work isolates and evaluates VPR purely at the retrieval stage, independently of any particular registration pipeline. Building on these results, future work will integrate the most promising VPR configurations into complete scene registration and SLAM pipelines and quantify their impact on pose accuracy and robustness while preserving runtime efficiency.

{
	\begin{spacing}{1.0}
		\normalsize
		\bibliography{references} 
	\end{spacing}
}

\end{document}